\documentclass{article}

\PassOptionsToPackage{square,numbers}{natbib}
\usepackage[preprint]{neurips_2019}
\usepackage{xcolor}

\usepackage{subfigure}
\usepackage{epsfig}
\usepackage[utf8]{inputenc} 
\usepackage[T1]{fontenc}  
\usepackage{hyperref}
\usepackage{color} 
\usepackage{url}      
\usepackage{booktabs}    
\usepackage{amsfonts}    
\usepackage{nicefrac}    
\usepackage{microtype}   
\usepackage{amsmath, amssymb, amsthm}
\usepackage{eqnarray}
\usepackage[makeroom]{cancel}
\newtheorem{theorem}{Theorem}
\newtheorem{lemma}{Lemma}
\usepackage[english]{babel}

\newcommand{\nn}{\nonumber}

\title{A Look at the Effect of Sample Design on Generalization through the Lens of Spectral Analysis
}

\author{%
  Bhavya Kailkhura\\
  Lawrence Livermore National Laboratories\\
  Livermore, CA 15213 \\
  \texttt{kailkhura1@llnl.gov} \\
  \And
  Jayaraman J. Thiagarajan\\
  Lawrence Livermore National Laboratories\\
  Livermore, CA 15213 \\
  \texttt{jjayaram@llnl.gov} \\
  \And
  Qunwei Li\\
  Lawrence Livermore National Laboratories\\
  Livermore, CA 15213 \\
  \texttt{li59@llnl.gov} \\
  \And
 Peer-Timo Bremer\\
  Lawrence Livermore National Laboratories\\
  Livermore, CA 15213 \\
  \texttt{bremer5@llnl.gov} \\
}

\begin{document}

\maketitle

\begin{abstract}
This paper provides a general framework to study the effect of sampling properties of training data on the generalization error of the learned machine learning (ML) models. Specifically, we propose a new spectral analysis of the generalization error, expressed in terms of the power spectra of the sampling pattern and the function involved. The framework is build in the Euclidean space using Fourier analysis and establishes a connection between some high dimensional geometric objects and optimal spectral form of different state-of-the-art sampling patterns.
Subsequently, we estimate the expected error bounds and convergence rate of different state-of-the-art sampling patterns, as the number of samples and dimensions increase. We make several observations about generalization error which are valid irrespective of the approximation scheme (or learning architecture) and training (or optimization) algorithms. Our result also sheds light on ways to formulate design principles for constructing optimal sampling methods for particular problems.
\end{abstract}

\section{Introduction}
Analyzing the generalization error of a learning algorithm is essential for estimating how well the generated hypothesis will apply to unknown test data. Traditionally, generalization error is analyzed based on the complexity of the function class, such as, the Vapnik-Chervonenkis (VC) dimension and the Rademacher complexity~\cite{gen1}, or properties of the learning algorithm, such as uniform stability~\cite{gen2}, and upper bounds on the error are derived. Recently, the authors in~\cite{gen3} showed that the mutual information between the collection of empirical risks of the available hypotheses and the final output of the algorithm can be used to analyze the generalization error in learning problems. In a similar information-theoretic setup, the authors in~\cite{gen4} proposed to bound generalization error using the total-variation distance.

Here, we are interested in studying generalization from the viewpoint of the sampler generating the training data. Sample design has been a long-standing research area in statistics, and a plethora of sampling solutions exist in the literature with a wide-range of assumptions and statistical guarantees; see~\citep{doe:review, Sampling:survey} for a detailed review of related methods. The properties of the sampling distribution directly control the expected convergence behavior of the generalization error, as sample size grows asymptotically. Consequently, designing an optimal sampler for a learning algorithm requires quantifying how the properties of the sampling distribution affects the generalization error. Unfortunately, existing theoretical tools for analyzing generalization error are not applicable for our purpose as they do not provide a direct connection to the sample properties, i.e.\ uniformity, randomness, etc. In this context, this paper addresses two important challenges: $1)$ identifying expressive metrics to quantify sample properties, and $2)$ bounding generalization error in terms of those tractable sample properties. 

Generically, a good sampling technique aims to cover the input space as uniformly as possible, in order to generate the so-called \textit{space-filling} experiment designs~\citep{sfd}. Since it is challenging to qualitatively evaluate the space-filling property, simple scalar metrics such as discrepancy~\cite{caflisch_1998} or geometric distances (maximin or minimax distance of a sample design~\citep{Schlomer:2011}) are utilized. However, recent studies have shown that these scalar metrics are not very descriptive, and when used as the design objective, often results in poor-quality samples~\cite{kailkhura2018spectral}. Furthermore, existing sampling distributions are not designed to specifically improve generalization error of learning algorithms. This is due to the lack of a principled framework for connecting sampling properties to generalization error. To address this challenge, we develop a novel spectral analysis framework to study generalization error, expressed in terms of the power spectra of sampling patterns as well as the function to be recovered. 

\noindent \textbf{Contributions and Findings}: First, we propose to adopt spectral analysis for characterizing the space-filling property of sampling patterns. More specifically, we use tools from statistical mechanics to connect the spectral properties of a sampling pattern with its spatial properties. Next, we develop an analysis framework for studying generalization error behavior of a learning algorithm through the lens of spectral properties of the sample design. Using this framework, for isotropic, homogeneous sampling patterns (i.e. we can use a radially averaged power spectrum), we derive best and worst-case generalization error bounds. 

While majority of existing literature on generalization error characterization based on sampling~\cite{gen2} have focused on uniform random sampling, the proposed analysis framework allows us to study the behavior of a large class of sample designs. In particular, we consider the blue noise~\cite{Kailkhura:2016:SBN, Heck:2013} and the Poisson disk sampling (PDS) distributions and obtain sampler-specific bounds (see Figure 1 in the supplementary material for examples of the distributions used). We characterize the gain due to blue noise and PDS samples over random sampler in a closed-form. 
This analysis further helps us to formulate design principles to construct optimal sampling methods to specific ML problems. Finally, we make interesting (counter-intuitive) observations on the convergence behavior of generalization error with increasing dimensions, and hence develop novel spectral metrics to obtain meaningful convergence results for different sampling patterns (included in the supplementary material).






\section{Preliminaries - Spectral Analysis for Sampling}
Fourier analysis is a classical approach for studying properties of sampling distributions. For example, the power spectral density (PSD) can be used to assess the quality of sampling distributions. Alternately, analyzing spatial characteristics of samples can also provide crucial insights. While such a spatial analysis has been traditionally carried out using heuristic measures for uniformity of sampling patterns, we adopt a more descriptive characterization.

\noindent \textbf{Power Spectral Density}: For a finite set of $N$ samples, $\{\mathbf{x}_j\}_{j=1}^N$, in a region with unit volume, the radially-averaged power spectral density describes how the signal power is distributed over frequencies. It is formally defined as
\begin{equation}
P(\mathbf{k}) = \frac{1}{N} |S(\mathbf{k})|^2 = \frac{1}{N} \sum_{j,\ell} e^{-2\pi i \mathbf{k.}(\mathbf{x}_{\ell} - \mathbf{x}_j)}, 
\end{equation}where $S(\mathbf{k})$ denotes the spectral coefficients. For isotropic distributions, we have $P(k)=P(|\mathbf{k}|)$.

\noindent \textbf{Pair Correlation Function}: A PCF describes the joint probability of having samples at two locations at the same time. It can be more precisely defined in terms of the intensity $\lambda$ and product density $\rho$ of a point process $X$~\cite{Oztireli:2012}. The intensity $\lambda(X)$ of $X$ is the average number of points in an infinitesimal volume around $X$. For isotropic point processes, this is a constant. Let $\{B_i\}$ denote the set of infinitesimal spheres around the points, and $\{dV_i\}$ denote the volume measures of $B_i$. The product density $P(\mathbf{x_1}, \cdots, \mathbf{x_N}) = \beta(\mathbf{x_1}, \cdots, \mathbf{x_N})dV_1\cdots dV_N$. In the isotropic case, for a pair of points, $\beta$ depends only on the distance between the points, hence one can write $\beta(\mathbf{x_i},\mathbf{x_j})=\beta(||\mathbf{x_i}-\mathbf{x_j}||)=\beta(r)$ and $P(r)=\beta(r)dxdy$. The PCF is then defined as $G(r)=\dfrac{\beta}{\lambda^2}.$

\noindent \textbf{Relating PCF and PSD via Fourier Transform}: The PSD and PCF of a point distribution are related via the Fourier transform as follows:
\begin{eqnarray*}
P(\mathbf{k})&=& 1+N F\left(G(\mathbf{r})-1\right) = 1+N \int_{\mathbb{R}^d} \left(G(\mathbf{r})-1\right)\exp(-2\pi i\mathbf{k.r}) d\mathbf{r}
\end{eqnarray*}
where $F(.)$ denotes the $d$-dimensional Fourier transform. 
Next, we establish a fundamental relationship between PSD and PCF for radially symmetric or isotropic distributions.
\begin{theorem}
\label{lemma1}
The pair correlation function and the power spectral density of radially symmetric function are related as follows:
\begin{equation*}
G(r)= 1+\frac{1}{N} r^{1-\frac{d}{2}} H_{\frac{d}{2}-1}\left(\rho^{\frac{d}{2}-1}(P(\rho)-1)\right)
\end{equation*}
where $H_v(f(x))=2\pi \int_{0}^{\infty}  x f(x) J_{v} (2\pi r x) dx$ is the Hankel transform. 
\end{theorem}
\begin{proof}
Please see supplementary material.
\end{proof}

\noindent \textbf{Realizability}: The two necessary mathematical conditions~\footnote{Whether or not these two conditions are not only necessary but also sufficient is still an open question (however, no counterexamples are known).} that a sampling pattern must satisfy to be realizable are: (a) its PSD must be non-negative, i.e., $P(\rho)\geq 0,\;\forall \rho$, and (b) its pair correlation function must be non-negative, i.e., $G(r)\geq 0,\;\forall r$.

\section{Risk Minimization using Monte Carlo Estimates}
\label{erm}
We consider the following general setup, which encompasses several supervised learning formulations. We consider two spaces of objects $X\in \mathbb{T}^d$ (toroidal unit cube $[0,1]^d$) and $Y\in \mathbb{R}$, and the goal is to learn a function $h:X\rightarrow Y$ (often called \textit{hypothesis}) which outputs $y\in Y$ for a given $x\in X$. We assume access to training data comprised of $N$ samples $S = \{(x_1,y_1),\cdots,(x_N,y_N)\}$ drawn i.i.d. from an unknown distribution $P(x,y)$. Supervised learning attempts to infer a hypothesis $h(.)$ that minimizes the population risk: 
\begin{equation}
  \label{pop_risk}
  R_P(h) \triangleq \mathbb{E}_{P(x,y)}[l(h(x),y)] = \int l(h(x),y) dP(x,y),
\end{equation}where $l(.,.)$ denotes the loss function.

\vspace{0.1in}

\noindent \textbf{Empirical Risk Minimization}: In general, the joint distribution $P(x,y)$ is unknown to the learning algorithm and hence the risk $R_P(h)$ cannot be computed. However, we often use an approximation, referred as \textit{empirical risk}, obtained by averaging the loss function on the training data:
\begin{equation}
  \label{emp_risk}
  {R}_S(h) \triangleq \frac{1}{N}\sum_{i=1}^N l(h(x_i),y_i)
\end{equation}
Note that the empirical risk ${R}_S(h)$ is a Monte Carlo (MC) estimate of the population risk $R_P(h)$. It also has a continuous form 
\begin{equation}
  \label{emp_risk_cont}
  R_S(h) \triangleq \frac{1}{N} \int_\mathbb{D} S(x) l(h(x),y) dx
\end{equation}
where $\mathbb{D}$ is the sampling domain, $S(x)$ is the sampling function, i.e., a sampling pattern rewritten as a random signal $S$ composed of $N$ Dirac functions located at sample positions $S(x)=\sum \delta(x-x_i)$ for $i=1,\cdots,N$. 

\vspace{0.1in}

\noindent \textbf{Generalization Error}: In ML and statistical learning theory, the performance of a supervised learning algorithm is measured by the generalization error, which measures how accurately an algorithm is able to predict outcome values for previously unseen data. More specifically, we adopt the following definition of generalization error:
\begin{equation}
  \label{gen_error}
  \text{gen}(h) \triangleq \mathbb{E}_S[(R_P(h)-R_S(h))^2]
\end{equation}
which is the expected difference between the population risk of the output hypothesis and its empirical risk on the training data. The generalization error also has an alternating form with a direct link to the statistical properties of the sampling pattern:
\begin{eqnarray*}
  \text{gen}(h) &\triangleq& \mathbb{E}_S[(R_P(h)-R_S(h))^2] = \mathbb{E}_S[(R_P(h)-\mathbb{E}(R_S(h))+\mathbb{E}(R_S(h))-R_S(h))^2]\\
  &=& bias^2+var(R_S(h))
\end{eqnarray*}
We consider sampling patterns which are homogeneous, i.e. statistical properties of a sample are invariant to translation over the domain $\mathbb{D}$. Homogeneous sampling patterns are unbiased in nature, thus, the generalization error arises only from the variance. Note that the variance analysis of Monte-Carlo integration has been considered in the literature~\cite{durand2011frequency, subr2013fourier, MCvariance} and we build upon these methods. However, similar analysis is of generalization error in an ML context has not been carried out yet. 

\section{Connecting Generalization Error with Spectral Properties of Samples}
\label{theory}
In this section, we will express generalization error (or variance) in terms of the power spectra of both $S$ and $l$. To this end, we use the Monte Carlo estimator of risk in the Fourier domain, and derive the variance in the Fourier domain by leveraging our homogeneity assumption on sampling patterns.

\subsection{Monte Carlo Estimator of Risk in the Spectral domain}
The MC estimator of the risk as given in equation \eqref{emp_risk_cont} can be characterized in the Fourier domain $\phi$ using the fact that dot-product of functions (the integral of the product) is equivalent to the dot-product of their Fourier coefficients. This allows to us to build the MC estimator of the risk as follows:
\begin{equation}
  \label{risk_fourier}
  R_S(h) \triangleq \frac{1}{N} \int_\mathbb{\phi} \mathcal{F}_S(\omega)\mathcal{F}_l(\omega)^\intercal d\omega
\end{equation}
where $\mathcal{F}_S$, $\mathcal{F}_l$ denote the Fourier transforms of the sampling function $S$ and the loss function $l$.

\subsection{Generalization Error via the Spectral Analysis}
We now use the spectral domain version of empirical risk to define generalization error:
\begin{eqnarray}
  \nonumber \text{gen}(h) &\triangleq& 
   {bias^2}+var(R_S(h)) = \left(\mathbb{E}(R_S(h))-R_P(h)\right)^2+\mathbb{E}(R_S(h)^2) - (\mathbb{E}(R_S(h)))^2\\
   &=& \left(\mathbb{E}(R_S(h))-R_P(h)\right)^2+ \dfrac{1}{N^2}\int_{\phi\times \phi}\mathbb{E}(\mathcal{F}_{S,l}(\omega,\omega'))d\omega d\omega' - (\mathbb{E}(R_S(h)))^2\label{gen_error_spec}
\end{eqnarray}
where
$\mathcal{F}_{S,l}(\omega,\omega') \triangleq \mathcal{F}_S(\omega)\cdot\mathcal{F}_l(\omega)^\intercal\cdot\mathcal{F}_S(\omega')^\intercal\cdot\mathcal{F}_l(\omega')$. 
Next, we provide an explicit closed-form relation of generalization error with the power spectra of both the sampling pattern and the loss function. To derive this relation, we first simplify \eqref{gen_error_spec} by restricting our analysis to homogeneous sampling patterns, which are unbiased. 

\begin{lemma}
The generalization error in terms of the power spectra of both the
sampling pattern and the loss function in the toroidal domain can be obtained as:
\begin{equation}
  \text{gen}(h) \triangleq \dfrac{1}{N}\int_{ \Theta} \mathbb{E}(\mathcal{P}_S(\omega)) \mathcal{P}_l(\omega) d\omega
\end{equation}
\end{lemma}
\begin{proof}
Please see supplementary material.
\end{proof}

If homogeneous sampling is isotropic (i.e., the power spectrum is radially symmetric), then the error can be computed from the radial mean power spectrum of the loss $\hat{\mathcal{P}_l}$ and the sampling pattern $\hat{\mathcal{P}_S}$. 
\begin{theorem}
The generalization error for isotropic homogeneous sampling patterns (in polar coordinates) is given by
\begin{equation}
\label{gen_rad}
  \text{gen}(h) \triangleq \dfrac{\mu(\mathcal{S}^{d-1})}{N}\int_{0}^{\infty} \rho^{d-1} \mathbb{E}(\hat{\mathcal{P}_S}(\rho)) \hat{\mathcal{P}_l}(\rho) d\rho,
\end{equation}
where $\mu(\mathcal{S}^{d-1})$ is the Lebesgue measure of a $(d-1)$-dimensional unit sphere in $\mathbb{R}^d$ given by $2\sqrt{\pi^d}/\Gamma(d/2)$ which is the surface area of the $(d-1)$-dimensional unit sphere.
\end{theorem}

\section{Best and Worst Case Generalization Error}
Using the proposed spectral analysis framework to predict generalization error requires us to explicitly know the power spectra of the loss function, which is usually unknown. Thus, we restrict our analysis to a particular class of integrable functions of the form $l(x)_{\mathcal{X}_\Omega}$ with $l(x)$ smooth and $\Omega$ a bounded domain with a smooth boundary ($\mathcal{X}_\Omega$ is the characteristic function of $\Omega$)~\cite{brandolini2001mean}. We consider a best-case function and a worst-case function, both from this class of functions to derive the error convergence rates, as the number of
samples $N$ and dimension $d$ grow. Note that the power spectra of sampling distributions are usually known in advance. We show that this information can be used in our framework to compute the generalization error bounds. Note, We perform our analysis following \eqref{gen_rad}, where the error is characterized by the radial mean power spectra of both sampling pattern and loss function.



\subsection{Best-Case Generalization Error}
We define our best-case function directly in the spectral domain with the
radial mean power spectrum profile $\hat{\mathcal{P}_l}(\rho)$ which is a constant $c_l$ for $(\rho<\rho_0)$, and zero elsewhere. The constant $c_l$ comes from the fact that the power spectrum is bounded. The best case error can be thus obtained from \eqref{gen_rad} as follows:

\begin{lemma}
The best-case generalization error for isotropic homogeneous sampling patterns (in polar coordinate) is given by
\begin{equation}
\label{gen_rad_best}
  \text{gen}(h) \leq \dfrac{\mu(\mathcal{S}^{d-1})}{N} c_l \int_{0}^{\rho_0} \rho^{d-1} \mathbb{E}(\hat{\mathcal{P}_S}(\rho)) d\rho.
\end{equation}
\end{lemma}

\subsection{Worst-Case Generalization Error}
For the worst-case, we consider our function to exhibit a radial
mean power spectrum which is $\hat{\mathcal{P}_l}(\rho)$ which is upper bounded by a constant $c_l$ for $(\rho<\rho_0)$, and $c_l'\rho^{-d-1}$ elsewhere, where $c_l$ and $c_l'$ are non-zero positive constants. This spectral profile has a decay rate $O(\rho^{-d-1})$ for $\rho>\rho_0$.

\begin{lemma}
The worst-case generalization error for isotropic homogeneous sampling patterns (in polar coordinate) is given by
\begin{equation}
\label{gen_rad_worst}
  \text{gen}(h) \leq \dfrac{\mu(\mathcal{S}^{d-1})}{N} c_l \int_{0}^{\rho_0} \rho^{d-1} \mathbb{E}(\hat{\mathcal{P}_S}(\rho)) d\rho+\dfrac{\mu(\mathcal{S}^{d-1})}{N} c_l' \int_{\rho_0}^{\infty} \rho^{-2} \mathbb{E}(\hat{\mathcal{P}_S}(\rho)) d\rho.
\end{equation}
\end{lemma}

\section{Sampler-Specific Generalization Error Bounds}
\label{sec:bounds}
In the previous section, we obtained the best and worst-case generalization error as a function of the sampling power spectrum $\mathbb{E}(\hat{\mathcal{P}_S}(\rho))$. In this section, we study the effects of different sampling distributions on the generalization error. 

\noindent \textbf{Random (or Poisson) Sampler}: This has a constant power spectrum since point samples are uncorrelated, i.e., $ \mathbb{E}(\hat{\mathcal{P}_S}(\rho)) = 1, \forall \rho$. For this spectral profile, the best-case generalization error can be obtained as:
\begin{eqnarray}
  \text{gen}_b(h) &\leq& \dfrac{\mu(\mathcal{S}^{d-1})}{N} c_l \int_{0}^{\rho_0} \rho^{d-1} d\rho = \dfrac{\mu c_l\rho_0^{d}}{N d} 
\end{eqnarray}
and the worst-case generalization error can be bounded as:
\begin{eqnarray}
  \text{gen}_w(h) &\leq& \dfrac{\mu(\mathcal{S}^{d-1})}{N} c_l \int_{0}^{\rho_0} \rho^{d-1} d\rho+\dfrac{\mu(\mathcal{S}^{d-1})}{N} c_l' \int_{\rho_0}^{\infty} \rho^{-2} d\rho\\
  &=& \dfrac{\mu c_l\rho_0^{d}}{N d} + \dfrac{\mu c_l'\rho_0^{-1}}{N} = \text{gen}_b(h)+ \dfrac{\mu c_l'\rho_0^{-1}}{N}
\end{eqnarray}

\noindent \textbf{Blue Noise Sampler}: Blue noise distributions are aimed at replacing visible aliasing
artifacts with incoherent noise, and its properties are typically
defined in the spectral domain. We consider the step blue noise pattern defined as follows: (a) the
spectrum should be close to zero for low frequencies, which indicates
the range of frequencies that can be recovered exactly; (b) the
spectrum should be a constant one for high frequencies, i.e. represent
uniform white noise, which reduces the risk of aliasing. The low frequency
band with minimal energy is referred to as the \textit{zero region}. Formally,
\begin{equation}
 P_S(\rho-\rho_z) = \left\{ \begin{array}{rll}
				0 & \mbox{if}\ \rho\leq \rho_z, \\
			  1 & \mbox{if}\ \rho>\rho_z.
				\end{array}\right. 
\label{steppsd}
\end{equation}The zero region $0\leq \rho \leq \rho_z$ indicates the range of frequencies that can be represented with no aliasing and the flat region $\rho> \rho_z$ guarantees that aliasing artifacts are mapped to broadband noise.

\begin{lemma}[\cite{Kailkhura:2016:SBN}]
\label{lemma3}
The pair correlation function of a Step blue noise sample of size $N$ in $d$ dimensions, for a given zero region $\rho_z$ is given by
\begin{equation}
\label{StepBN_pcf}
G(r) = 1-\frac{1}{N} \left(\frac{\rho_z}{r}\right)^{\frac{d}{2}} J_{\frac{d}{2}}(2\pi \rho_z r).
\end{equation}
\end{lemma}

Using Lemma~\ref{lemma3}, we can pose an
optimization problem for determining the maximum achievable zero
region $\rho_z$, that does not violate realizability conditions, for a
given sample budget $N$. 

\begin{lemma}
\label{lemma2}
The maximum achievable zero region using $N$ Step blue noise samples in
$d$ dimensions is equal to inverse of the $d$-th root of the volume of
a $d$-dimensional hyper-sphere with radius
$1/\sqrt[\leftroot{-2}\uproot{2}d]{N}$,
\begin{equation*}
\rho_z^* = \sqrt[\leftroot{-2}\uproot{2}d]{\dfrac{N \Gamma\left(1+\frac{d}{2}\right)}{\pi^{d/2}}}
\end{equation*}
where $\Gamma(.)$ is the gamma function. Equivalently, we can determine the minimum number of samples needed to construct a step blue noise pattern, $N = \frac{\pi^{d/2} \rho_z^d}{\Gamma(1+d/2)}$.

\end{lemma}
\begin{proof}
Please refer to the supplementary material.
\end{proof}




For this spectral profile, the best-case generalization error can be obtained as:
\begin{eqnarray}
  \text{gen}_b(h) &\leq& \dfrac{\mu(\mathcal{S}^{d-1})}{N} c_l \int_{0}^{\rho_0} \rho^{d-1} P_S(\rho-\rho_z^*) d\rho.
  \end{eqnarray}
Note that, when $\rho_0\leq \rho_z^*$ the best-case generalization error $\text{gen}_b(h)=0$, and when $\rho_0> \rho_z^*$, we have
 \begin{eqnarray}
  \text{gen}_b(h) &\leq& \dfrac{\mu(\mathcal{S}^{d-1})}{N} c_l \int_{\rho_z^*}^{\rho_0} \rho^{d-1} d\rho = \dfrac{\mu c_l}{N}\left[\dfrac{\rho_0^d-\rho_z^{*d}}{d}\right] \\ &=&\text{gen}_b^{\text{random}}(h)-\dfrac{\mu c_l \rho_z^{*d}}{N d} = \text{gen}_b^{\text{random}}(h)-\dfrac{\mu c_l \Gamma\left(1+d/2\right)}{d \pi^{d/2}}
  \end{eqnarray}

The worst-case generalization error can be obtained as:
\begin{eqnarray}
  \text{gen}_w(h) &\leq& \dfrac{\mu(\mathcal{S}^{d-1})}{N} c_l \int_{0}^{\rho_0} \rho^{d-1} P_S(\rho-\rho_z^*) d\rho+\dfrac{\mu(\mathcal{S}^{d-1})}{N} c_l' \int_{\rho_0}^{\infty} \rho^{-2} P_S(\rho-\rho_z^*) d\rho
\end{eqnarray}
{Note that, when $\rho_0\leq \rho_z^*$ the worst-case generalization error $\text{gen}_w(h)=\dfrac{\mu c_l'}{N \rho_z^*}$,} and when $\rho_0> \rho_z^*$,
\begin{eqnarray}
  \text{gen}_w(h) &\leq& \dfrac{\mu(\mathcal{S}^{d-1})}{N} c_l \int_{\rho_z^*}^{\rho_0} \rho^{d-1} d\rho+\dfrac{\mu(\mathcal{S}^{d-1})}{N} c_l' \int_{\rho_0}^{\infty} \rho^{-2} d\rho\\
  &=& \dfrac{\mu c_l}{N}\left[\dfrac{\rho_0^d-\rho_z^{*d}}{d}\right]+\dfrac{\mu c_l' \rho_0^{-1}}{N}\\
  &=& \text{gen}_b(h)+\dfrac{\mu c_l' \rho_0^{-1}}{N} = \text{gen}_w^{\text{random}}(h)-\dfrac{\mu c_l \Gamma\left(1+d/2\right)}{d \pi^{d/2}}
  \end{eqnarray}

\noindent \textbf{Poisson Disk Sampler}:
Without any prior knowledge of the function $f$ of interest, a reasonable objective for sampling is that the samples should be random to provide an equal chance of finding
features of interest. However, to avoid sampling only parts of the parameter space, a second objective is required to cover the space
in $\mathcal{D}$ uniformly. Poisson Disk Sampling (PDS) pattern are designed to achieve these objectives. In particular, the \textit{step PCF} sampling pattern is a set of samples that are distributed according to a uniform probability distribution (Objective 1: Randomness) but no two samples are closer
than a given minimum distance $r_{min}$ (Objective 2: Coverage). Formally,
\begin{equation}
 G_S(r-r_{min}) = \left\{ \begin{array}{rll}
				0 & \mbox{if}\ r\leq r_{min}, \\
			  1 & \mbox{if}\ r>r_{min}.
				\end{array}\right. 
\label{steppsd}
\end{equation}The PDS can also be defined in the spectral domain as follows:
\begin{lemma}[\cite{kailkhura2018spectral}]
\label{lemma6}
The power spectra of an ideal Poisson disk sampling pattern of size $N$ in $d$ dimensions, for a given $r_{min}$ is given by
\begin{equation}
\label{Step_psd}
P_S(\rho-r_{min}) = 1-N\left(\dfrac{2\pi r_{min}}{\rho}\right)^{d/2} J_{d/2}(\rho r_{min}),
\end{equation}
where $J_{d/2}(.)$ is the Bessel function of order $d/2$.
\end{lemma}

Similar to the previous case, we can determining the maximum achievable $r_{min}$, that does not violate realizability conditions, for a
given sample budget $N$. 
\vspace{0.15in}
\begin{lemma}
\label{lemma7}
The maximum achievable $r_{min}$ using $N$ Step PCF samples in $d$ dimensions is equal to inverse of the $d$-th root of the volume of
a $d$-dimensional hyper-sphere with radius
$\sqrt[\leftroot{-2}\uproot{2}d]{N}$,
\begin{equation*}
r_{min}^* = \sqrt[\leftroot{-2}\uproot{2}d]{\dfrac{ \Gamma\left(1+\frac{d}{2}\right)}{\pi^{d/2}N}}
\end{equation*}
where $\Gamma(.)$ is the gamma function. Equivalently, we can also determine the minimum $N$ required to achieve a given $r_{min}$, $N = \frac{\Gamma(1+d/2)}{\pi^{d/2} r_{min}^d}$.
\end{lemma}

\vspace{0.15in}



For PDS sampling, the best-case generalization error can be obtained as:
\begin{eqnarray}
  \text{gen}_b(h) &\leq& \dfrac{\mu(\mathcal{S}^{d-1})}{N} c_l \int_{0}^{\rho_0} \rho^{d-1} P_S(\rho-r_{min}^*) d\rho \nonumber\\
  &=& \dfrac{\mu c_l \rho_0^d}{N d}-\mu c_l(2\pi r_{min}^*)^{d/2} \int_{0}^{\rho_0} {\rho^{\frac{d}{2}-1} J_{d/2}(\rho r_{min}^*)}d\rho \nonumber\\
  &=& \text{gen}_b^{\text{random}}(h)-\mu c_l(2\pi r_{min}^*)^{d/2} \int_{0}^{\rho_0} {\rho^{\frac{d}{2}-1} J_{d/2}(\rho r_{min}^*)}d\rho \nonumber\\
  &=& \text{gen}_b^{\text{random}}(h)-\mu c_l(2\pi)^{d/2} r_{min}^* \int_{0}^{\rho_0} {(\rho r_{min}^*)^{\frac{d}{2}-1} J_{d/2}(\rho r_{min}^*)}d\rho\label{pds_b}
  \end{eqnarray}

The worst-case generalization error can be obtained as:
\begin{eqnarray}
  \text{gen}_w(h) &\leq& \text{gen}_b(h)+\dfrac{\mu(\mathcal{S}^{d-1})}{N} c_l' \int_{\rho_0}^{\infty} \rho^{-2} P_S(\rho-r_{min}^*) d\rho \nonumber\\
  &=& \text{gen}_b(h)+\dfrac{\mu}{N} c_l' \int_{\rho_0}^{\infty} \rho^{-2} d\rho-\mu c_l'(2\pi r_{min}^*)^{d/2} \int_{\rho_0}^{\infty} \rho^{-\frac{d}{2}-2} J_{d/2}(\rho r_{min}^*) d\rho\nonumber \\
  &=& \text{gen}_b(h)+\dfrac{\mu c_l' \rho_0^{-1}}{N} -\mu c_l'(2\pi r_{min}^*)^{d/2} \int_{\rho_0}^{\infty} \rho^{-\frac{d}{2}-2} J_{d/2}(\rho r_{min}^*) d\rho \nonumber\\
    &=& \text{gen}_b(h)+\dfrac{\mu c_l' \rho_0^{-1}}{N} - \mu c_l'(2\pi)^{d/2}r_{min}^{*{d+2}} \int_{\rho_0}^{\infty} (\rho r_{min}^*)^{-\frac{d}{2}-2} J_{d/2}(\rho r_{min}^*) d\rho\label{pds_w}
\end{eqnarray}

These integrals are complicated to compute and it is non-trivial to get clean and general bounds. Further simplifications under certain simplistic assumptions are provided in the supplementary material. 


\section{Convergence Analysis of Generalization Error}
Next, we analyze the convergence of error with blue noise and PDS sampling patterns with sample size $N$. This analysis will shed light into design principles for constructing sampling patterns.
\begin{figure*}[t]
\label{pds_rate}
\centering
\subfigure[]{
\includegraphics[%
 width=0.46\textwidth,clip=true]{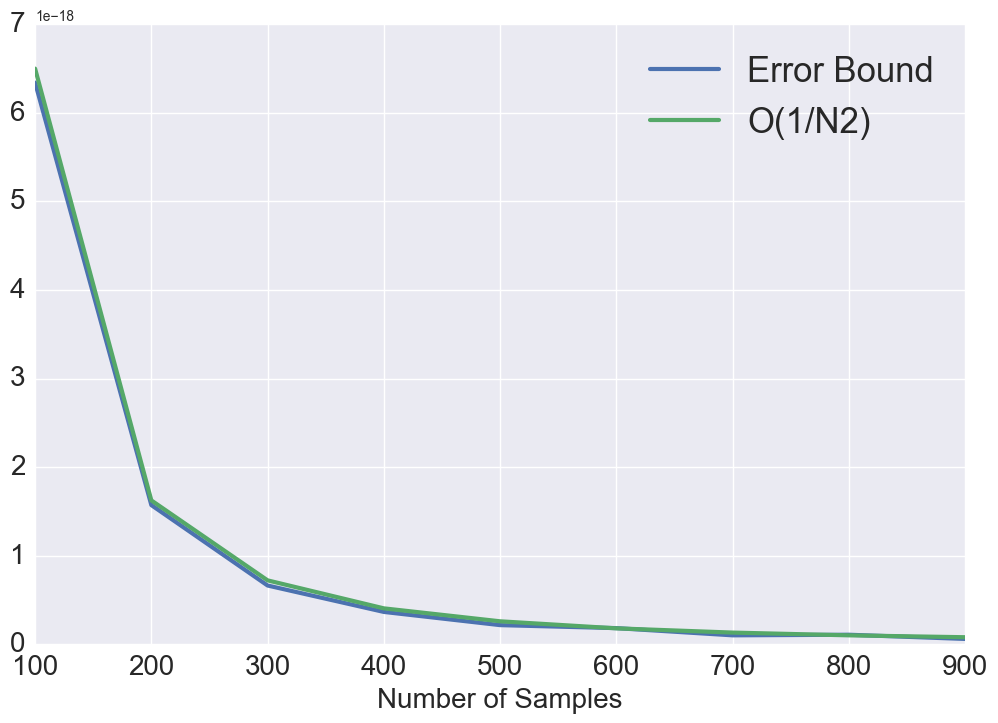}
\label{eb_n} }
\subfigure[]{
\includegraphics[%
 width=0.48\textwidth,clip=true]{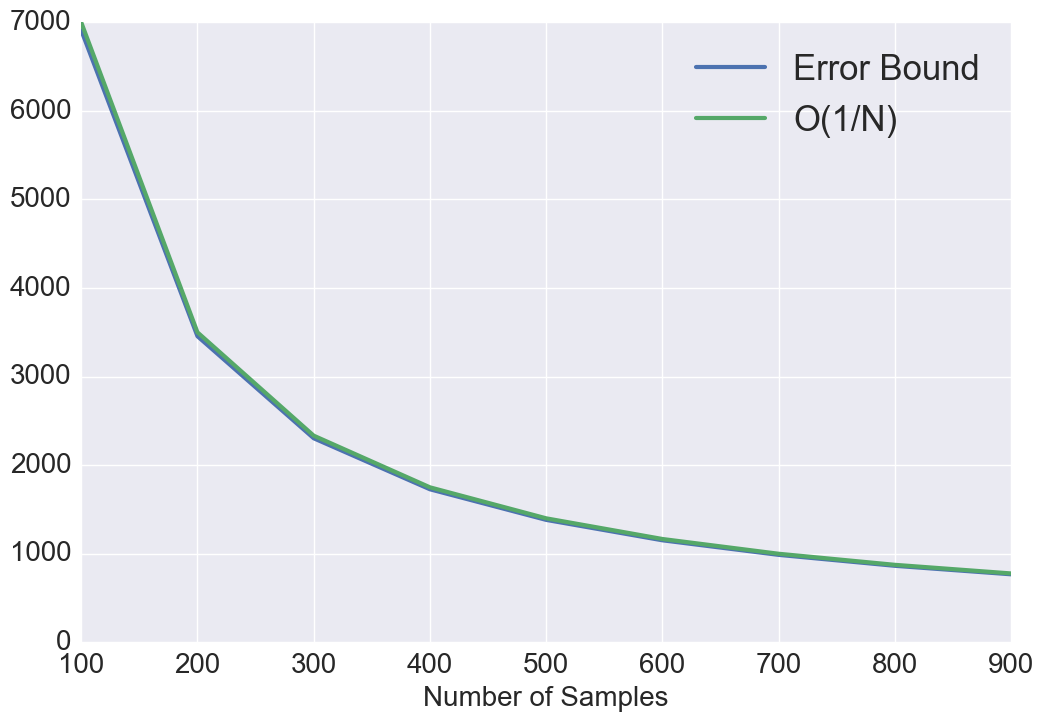}
\label{ew_n} }
\caption{Poisson Disk Sampling: (a) Best case convergence rate for $d=2$, (b) Worst case convergence rate for $d=2$. We use the parameters $c_l =10^8, c_l' = 1.1$ and $\rho_0 =10^{-4}$.}
\end{figure*}

\subsection{Analysis with Sample Size}
For random sampling patterns, both the best and the worst case generalization errors converge as $O\left(\frac{1}{N}\right)$. For blue noise sampling, if best case functions/signals are bandwidth-limited with $\rho_0\leq \rho_z^*$, then it can be perfectly recovered. However, when $\rho_0> \rho_z^*$, the convergence is at the rate $O\left(\frac{1}{N}\right)$, which is the same as random sampling. For worst case functions, the error converges as $O\left(\frac{1}{N \sqrt[\leftroot{-2}\uproot{2}d]{N}}\right)$ when $\rho_0\leq \rho_z^*$ and as $O\left(\frac{1}{N}\right)$ when $\rho_0 > \rho_z^*$. This provides a theoretical justification of designing a blue noise sampling pattern with a large zero-region $\rho_z$ for better performance. Note that the convergence rate analysis of Poisson disk sampling is not straightforward due to the involvement of Bessel functions under the integral in \eqref{pds_b} and \eqref{pds_w}. Hence, we numerically analyze the convergence for PDS pattern. As showed in Fig.~\ref{pds_rate}, We observe that the best case convergence rate approximately behaves as $O\left(\frac{1}{N \sqrt[\leftroot{-2}\uproot{2}d]{N}^b}\right)$ with $b\geq 1$ and the worst case convergence behaves as $O\left(\frac{1}{N}\right)$.

\subsection{Some Guidelines for Sample Design}
Results from the convergence analysis suggest that an ideal sampling power spectrum must attain zero values in the low frequency regime. Note that the realizability conditions severely limit the range of realizable power spectra and hence in practice, this results in blue noise patterns with very small $\rho_z$. Consequently, when the function is complex with $\rho_0>\rho_z^*$, a blue noise sample design behaves similar to a random design, $O(1/N)$. On the other hand, Poisson disk samples have a better error convergence rate even for complex functions compared to blue noise patterns. However, when $\rho_0\leq \rho_z^*$, blue noise pattern is ideal. This suggests that an ideal sampling pattern should trade-off the two paradigms by developing a sampling pattern that simultaneously carries the blue noise and PDS
properties. 

In many practical scenarios, it is possible to use information acquired from previous observations to improve the sampling process. As more samples are obtained, one can learn how to improve the sampling process by deciding where to sample next. These sampling feedback techniques are more generally known as adaptive sampling in the statistics literature. Our analysis provides a novel way to quantify the value of sample in terms of generalization error. A natural extension of our work is towards building importance sampling techniques, guided by spectral properties.


\section{Conclusions}
In this paper, we develop a framework to study the interplay between the sampling properties and the generalization error. We expressed generalization error in terms of power spectra of sampling pattern and the function of interest. We also analyzed the generalization error of some state-of-the-art sampling pattern and quantified their gain over random sampler in a closed-form. Finally, we provided some design guidelines for constructing optimal sampling patterns for a given problem. There are still many interesting questions that remain to be explored in the future work such as an analysis of the generalization error for cases where data comes from non-linear manifolds. Note that some analytical methodologies used in this paper are certainly exploitable for studying the effect of sample design on generalization error in different manifolds. Other questions such as PSD/PCF parameterizations for other variants of space-filling designs, adaptive and importance sampling, and optimization approaches to synthesize them can also be investigated.

\section{Acknowledgments}
This work was performed under the auspices of the U.S. Department of Energy by Lawrence Livermore National Laboratory under Contract DE-AC52-07NA27344.

 {\small{
\bibliographystyle{IEEEbib}
\bibliography{Ref}
}}

 \newpage

 \section{Appendix}
\section{Description of Sampling Distribution Families}
\begin{figure}[h]
    \centering
    \includegraphics[width = 0.8\linewidth]{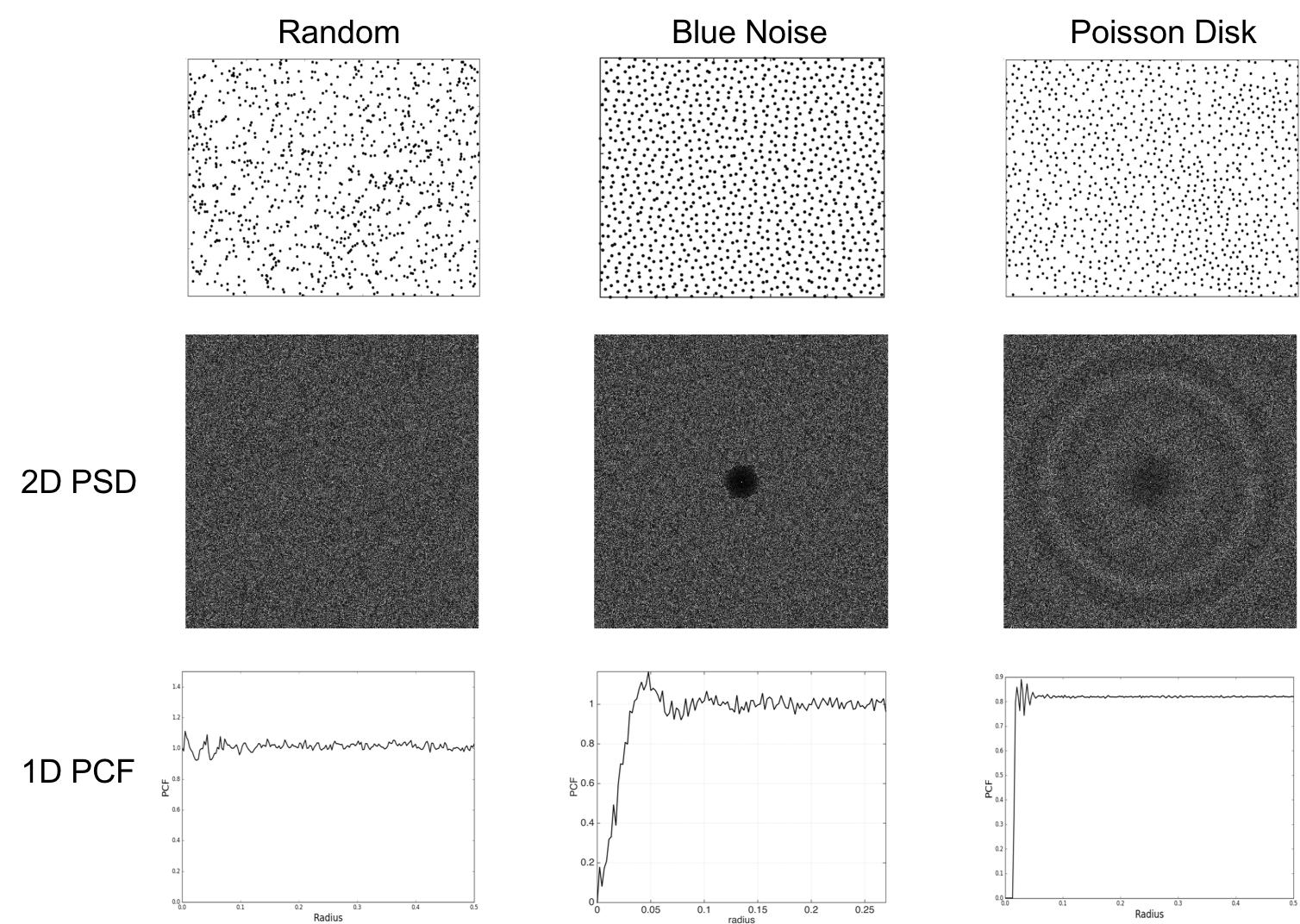}
    \caption{Sampling distributions along with their spatial/spectral properties considered in our analysis.}
    \label{fig:sampler}
\end{figure}

In this paper, we consider three different families of sampling patterns for our generalization error analysis, namely random, blue noise and Poisson disk sampling. Figure \ref{fig:sampler} illustrates the point distributions along with their spectral/spatial properties for $d = 2$ and $N = 1000$. Note that, we show the 2D PSD here, though our analysis assumes isotropic distributions and hence uses radially averaged 1D-PSD.

\section{Proof for Lemma 1 from the main paper}
The proof follows from \cite{MCvariance} and provided here for completeness. 

Let us denote the Fourier domain without the DC peak frequency as $\Theta$. Since homogeneous sampling patterns have statistical properties that are invariant to translation, it is equivalent to studying the error due to the translated version of each realization, with the average computed over all translations. Formally, we can treat the torus as the group of translations, so that $\tau(S)$ denotes the translation of $S$ by an element $\tau \in \mathcal{T}^d$. Then, averaging equation (7) over all translations of $S$, we get:

\begin{eqnarray}
  \text{gen}(h) &\triangleq& \dfrac{1}{N^2}\int_{\mathcal{T}^d \times \Theta\times \Theta}\mathbb{E}(\mathcal{F}_{\mathcal{T}(S),l}(\omega,\omega'))d\omega d\omega'd\tau,\\
  &=& \dfrac{1}{N^2}\int_{\mathcal{T}^d \times \Theta\times \Theta}\mathbb{E}(\mathcal{F}_{S,l}(\omega,\omega'))\exp^{i2\pi \tau\cdot(\omega'-\omega)} d\omega d\omega'd\tau,
  \label{gen_error_homo}
\end{eqnarray}where the exponential arises from the translation of the sampling pattern by a vector $\tau$ in the Fourier domain. 
When $\omega \neq \omega'$, the
integral of the exponential part equals zero, so that only the case $\omega=\omega'$ contributes to the variance. Hence, we can remove one
integral over $\Theta$ and obtain
\begin{eqnarray}
  \text{gen}(h) &\triangleq& \dfrac{1}{N^2}\int_{ \Theta}\mathbb{E}(\mathcal{F}_{S,l}(\omega,\omega)) \int_{\mathcal{T}^d} d\omega d\tau\\
  &=& \dfrac{1}{N^2}\int_{ \Theta}\mathbb{E}(\|\mathcal{F}_{S,l}(\omega,\omega)\|^2) d\omega
\end{eqnarray}
Finally, denoting the power spectrum of the loss by $\mathcal{P}_l$ and the power spectrum of the sampling pattern normalized by $N$ as $\mathcal{P}_S$, and leveraging the fact that $\|\mathcal{F}_{S,l}(\omega,\omega)\|^2=\|\mathcal{F}_{S}(\omega)\|^2\cdot \|\mathcal{F}_{l}(\omega)\|^2$,
\begin{equation}
  \text{gen}(h) \triangleq \dfrac{1}{N}\int_{ \Theta} \mathbb{E}(\mathcal{P}_S(\omega)) \mathcal{P}_l(\omega) d\omega
\end{equation}This provides the expression for the generalization error in terms of the power spectra of both the sampling pattern and the loss function in the toroidal domain.

\section{Proof of Theorem 1 from the main paper}
\label{app1}
We know that the PSD and PCF of a point distribution are related via the Fourier transform as follows:
\begin{eqnarray*}
P(\mathbf{k})&=& 1+\rho F\left(G(\mathbf{r})-1\right)\\
&=& 1+N \int_{\mathbb{R}^d} \left(G(\mathbf{r})-1\right)\exp(-2\pi i\mathbf{k.r}) d\mathbf{r}
\end{eqnarray*}
where $F(.)$ denotes the $d$-dimensional Fourier transform. Using symmetry of the Fourier transform, we have
\begin{eqnarray*}
G(\mathbf{r})&=& 1+\frac{1}{N} F\left(P(\mathbf{k})-1\right).
\end{eqnarray*}
Next, we use polar coordinates with the $z$ axis along $\mathbf{k}$, so that $\mathbf{k.r} = \rho r \cos \theta$ where $\rho = |\mathbf{k}|$ and $r = |\mathbf{r}|$. For radially symmetric PCF, we have $G(\mathbf{r})=G(r)$ and the above relationship can be rewritten as 

\begin{eqnarray*}
G(r)&=& 1+\frac{1}{N} \int_{0}^{\infty} \int_{0}^{\pi} \exp\left(-2\pi i \rho r \cos (\theta)\right)\\
&&\qquad (P(\rho)-1) \omega \sin (\theta)^{d-2} d\theta r^{d-1} \; d\rho
\end{eqnarray*}
where $\omega$ is the area of unit sphere in $(d-1)$ dimension. Next, using the identity involving bessel function of order $v$, i.e., 
\begin{eqnarray*}
J_{v}(2\pi t)&=& \frac{(2\pi t)^{v}}{(2\pi)^{v+1}}\int_{0}^{\pi} \exp\left(-2\pi i t \cos (\theta)\right)\\
&&\qquad \omega \sin (\theta)^{2v} \;d\theta, 
\end{eqnarray*}
we obtain 
\begin{eqnarray*}
G(r)&=& 1+\frac{1}{N} r^{1-\frac{d}{2}} 2\pi \int_{0}^{\infty} \rho^{\frac{d}{2}-1} J_{\frac{d}{2}-1} (2\pi \rho r) \rho (P(\rho)-1)\; dk.
\end{eqnarray*}

\section{Proof of Lemma 5 from the main paper}
\label{app2}
Note that, for a Step blue noise configuration to be realizable, it is sufficient to show that the corresponding PCF is non-negative. Thus, we have 
\begin{eqnarray*}
&& G(r) \geq 0\\
&\Leftrightarrow& 1\geq \frac{1}{N} \left(\frac{\rho_z}{r}\right)^{\frac{d}{2}} J_{\frac{d}{2}}(2\pi \rho_z r)\\
&\Leftrightarrow& 1\geq \frac{1}{N} (\rho_z \sqrt{2 \pi})^d \frac{J_{\frac{d}{2}}(2\pi \rho_z r)}{(2\pi \rho_z r)^{\frac{d}{2}}}\\
&\Leftrightarrow& 1\geq \frac{1}{N} \frac{(\rho_z \sqrt{2 \pi})^d}{2^{\frac{d}{2}}\Gamma\left(1+\frac{d}{2}\right)} \\
&\Leftrightarrow& \rho_z \leq \left(\dfrac{N \Gamma\left(1+\frac{d}{2}\right)}{\pi^{d/2}}\right)^{1/d}.
\end{eqnarray*}
 In the last inequality, we have used the following approximation
 $$J_v(x) = \frac{(x/2)^v}{\Gamma(1+v)}.$$
 
 \section{Generalization Error Bounds for Poisson Disk Sampling Patterns}
 
\subsection*{Best Case}
 \begin{align}
   \text{gen}_b(h) &\le \text{gen}_b^{\text{random}}(h)-\mu c_l(2\pi)^{d/2} r_{min}^* \int_{0}^{\rho_0} {(\rho r_{min}^*)^{\frac{d}{2}-1} J_{d/2}(\rho r_{min}^*)}d\rho \nn \\
   & = \text{gen}_b^{\text{random}}(h)-\mu c_l(2\pi)^{d/2} \int_{0}^{r_{min}^*\rho_0} {\rho^{\frac{d}{2}-1} J_{d/2}(\rho)}d\rho\nn \\
  & \le \text{gen}_b^{\text{random}}(h) - \frac{\mu c_l(2\pi)^{d/2} 2^{\frac{-d}{2}}{(\rho r_{min}^*)}^d }{d\Gamma (1+\frac{d}{2}) }\left(1-\frac{1}{8}\frac{d{(\rho r_{min}^*)}^2 }{(1+\frac{d}{2})^2}\right)\nn \\
   &= \frac{\mu c_l \Gamma ^{\frac{2}{d}}(1+\frac{d}{2})\rho_0^{2+d}}{ 8\pi(1+\frac{d}{2})^2} \frac{1}{N^{1+\frac{2}{d}}}
 \end{align}
 The second inequality above is based on the series form of the hypergeometric function and the assumption that $N$ is a large number.


 \begin{figure*}[t]
\centering
\subfigure[]{
\includegraphics[%
 width=0.45\textwidth,clip=true]{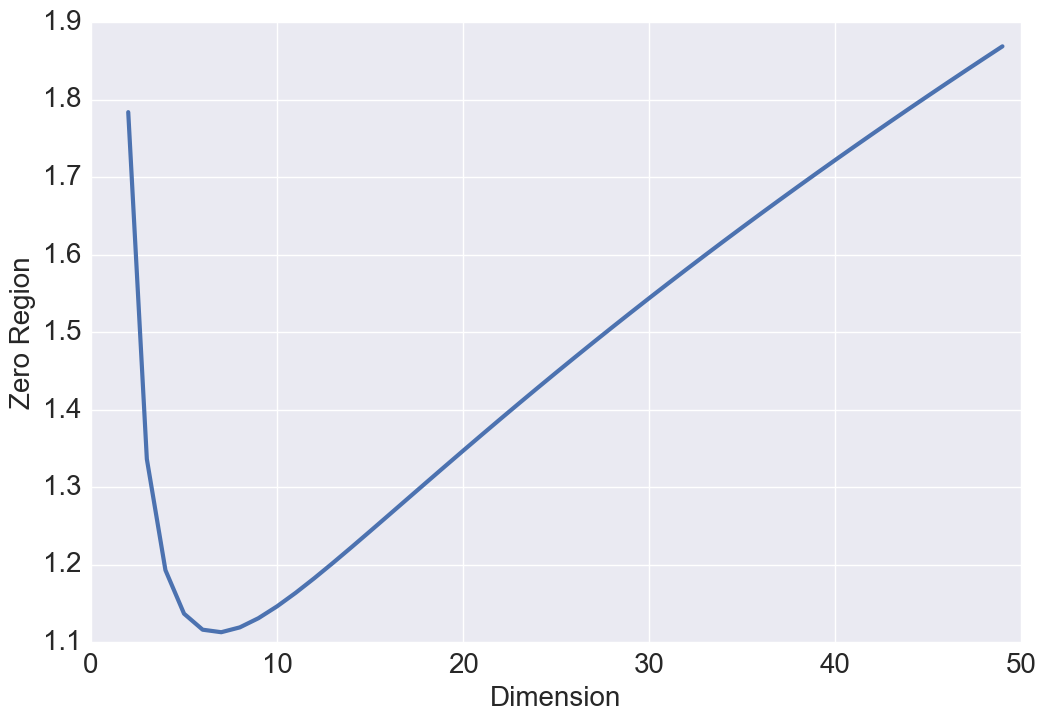}
\label{rho_d}}
\subfigure[]{
\includegraphics[%
 width=0.45\textwidth,clip=true]{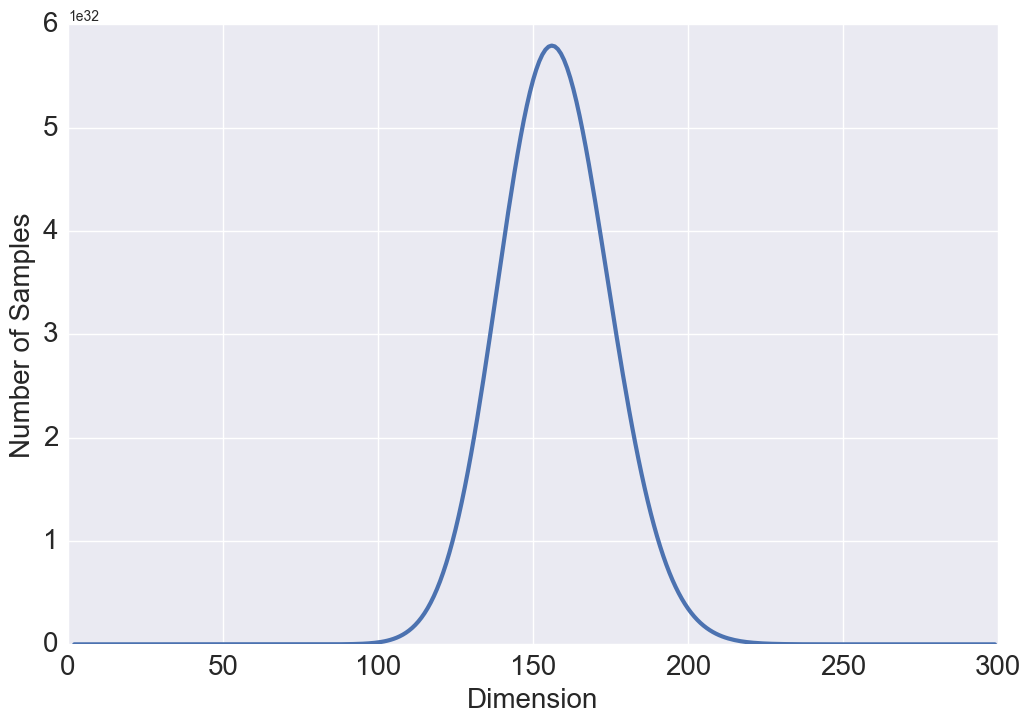}
\label{Nz_d}}
\subfigure[]{
\includegraphics[%
 width=0.45\textwidth,clip=true]{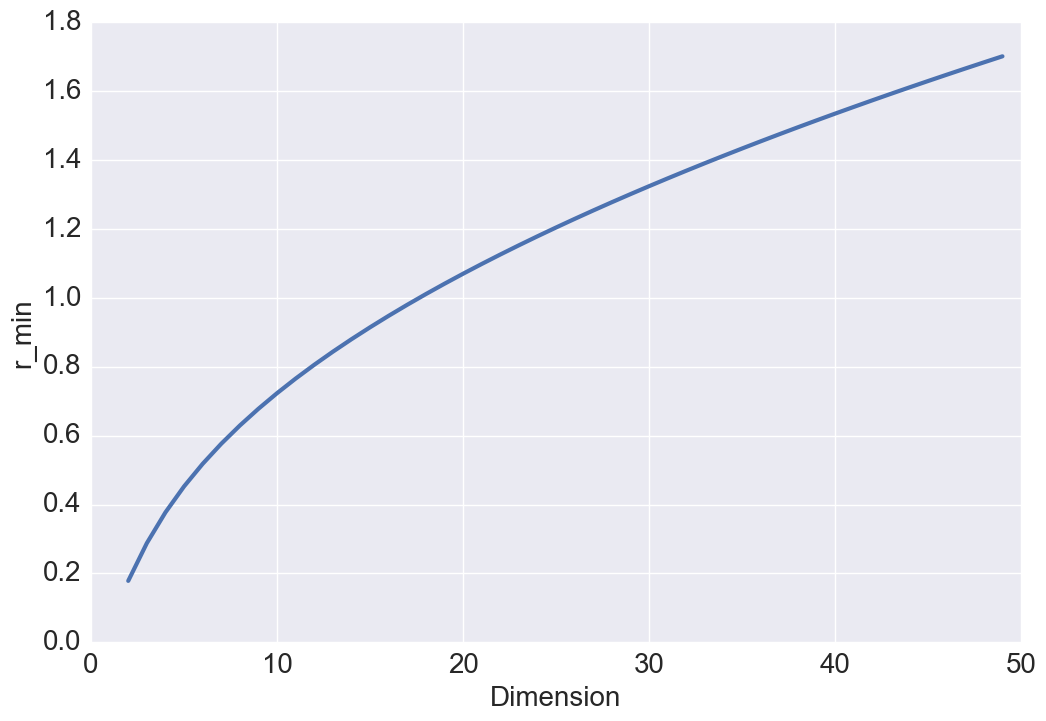}
\label{rmin_d} }
\subfigure[]{
\includegraphics[%
 width=0.45\textwidth,clip=true]{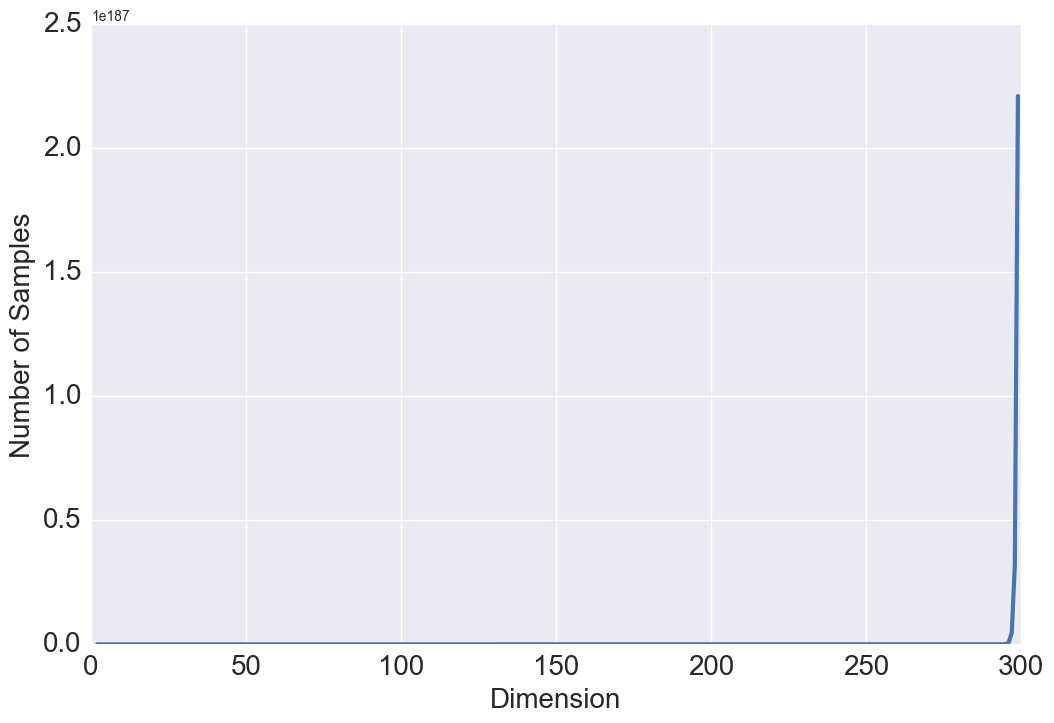}
\label{Nmin_d} }
\caption{(a) Maximum achievable $\rho_z$ with $N = 10$ samples of blue noise at varying dimensions $d$, (b) Minimum number of sampling points needed to achieve $\rho_z = 5$ at varying dimensions $d$, (c) Maximum achievable $r_{min}$ with $N = 10$ samples of PDS at varying dimensions $d$, (d) Minimum number of sampling points needed to achieve $r_{min} = 1$ at varying dimensions $d$.}
\end{figure*}
 
 \section{Convergence Analysis of Generalization Error with Dimensions}
In this section, we report some interesting observations when analyzing generalization error with increasing dimensions. We study the limiting behavior of $\rho_z^*$ and $r_{min}^*$ as $d$ approaches infinity. We show that the analysis with conventional metrics to characterize the zero region, i.e., the range of frequencies that can be represented with no aliasing, provides some rather counter-intuitive results.

\begin{lemma}
 As the dimension $d$ approaches infinity, the maximum achievable zero region for blue noise sampling, with a fixed $N$, goes to infinity, i.e.,
$ \lim_{d\rightarrow \infty} \rho_z^* = \infty$
and, the minimum number of samples needed to achieve a zero region $\rho_z$ approaches zero, i.e., 
$ \lim_{d\rightarrow \infty} N = 0.$
\end{lemma}Intuitively, with gowing $d$, one might expect $\rho_z^* \rightarrow 0$ and $N \rightarrow \infty$. To better understand this result, we study the relationship between these two quantities and the volume of a hyper-sphere. One of the surprising facts about a sphere in high dimensions is that as the dimension increases, the volume of the sphere goes to zero which justifies the above results. Our intuitions about space are formed in two or three dimensions and often do not hold in high dimensions. A more surprising fact is that $\rho_z^*$ and $N$ are not monotonic functions with respect to $d$ (see Figure~\ref{rho_d} and~\ref{Nz_d}). Either a steady increase or a steady decrease seems more plausible than having these two quantities grow for a while, then reach a peak at some finite value of $d$, and thereafter decline. This behavior has also been observed in high dimensional geometry while analyzing the volume of a hypersphere, however, no physical interpretation or intuition currently exists for this open research problem~\cite{Hayes}.

Similarly, we study the asymptotic behavior of the maximum achievable $r_{min}$ for a fixed sample budget, and equivalently the minimum number of samples required to achieve a PDS with a given $r_{min}$, as the dimension grows to infinity.
\begin{lemma}
 As the dimension $d$ approaches infinity, the maximum achievable $r_{min}$ for PDS sampling pattern, with a fixed number of samples, goes to infinity, i.e.,
 $\lim_{d\rightarrow \infty} r_{min}^* = \infty$
and, the minimum number of samples needed to achieve a $r_{min}$ also approaches infinity, i.e., 
$\lim_{d\rightarrow \infty} N = \infty.$
\end{lemma}The results in the lemma above are reasonable, since the space is growing exponentially fast. 

\subsection{Analysis with Proposed Metrics}
Analysis with the metrics $\rho_z^*$ and $r_{min}^*$, which are based on the amplitude of the
frequency vector, i.e., $\mathbf{k}$, to characterize the zero region, leads to inconsistent results in high dimensions. We argue that comparing $\rho_z^*$ and $r_{min}^*$ across different dimensions is not accurate, and these inconsistent results are a byproduct of the improper comparisons. Note that, each $d$-dimensional space is comprised of a different range of frequency components, and comparing the magnitude of the frequency vector directly across dimensions is questionable. In particular, for a valid comparison of volumes across dimensions, we propose to measure them in terms of a standard volume in that dimension, i.e., unit hypercube or the measure polytope, which has a volume of $1$ in all dimensions. Further, as the dimension $d$ increases, the maximum possible distance between two points in a hypercube grows as $\sqrt{d}$. Consequently, to have same scale across dimensions, we normalize the radius of the hypersphere by the factor $\sqrt{d}$. In summary, we introduce the \textit{relative zero region}, i.e., $\hat{\rho_z}^* = \rho_z^*/\sqrt{d}$ ($\hat{r}_{min}^* = r_{min}^*/\sqrt{d}$) for meaningful convergence analysis across dimensions.

\begin{lemma}
\label{rho_r}
As dimension $d$ approaches infinity, the maximum achievable relative $\rho_z^*$ converges to a constant, i.e.,
$$\lim_{d\rightarrow \infty} \hat{\rho}_z^* = \frac{1}{\sqrt{2\pi e}} $$
and, the minimum number of blue noise samples needed to achieve $\hat{\rho}_z^*$ goes to infinity.
\end{lemma}
\begin{proof}
To prove the first identity, note that $\dfrac{\rho_z^*}{\sqrt(d)} = \dfrac{\sqrt[\leftroot{-2}\uproot{2}d]{N}}{\sqrt{\pi d}} \sqrt[\leftroot{-2}\uproot{2}d]{\Gamma\left(1+\frac{d}{2}\right)}$ and invoke Stirling's approximation, i.e., $\Gamma(1+m) = \left(\frac{m}{e}\right)^m \sqrt{2\pi m}$. Now, the required result can be obtained by letting $d$ approach infinity. 
The second identity can be proved in a similar manner. 
\end{proof}

Similarly, we study the asymptotic behavior of $\hat{r}_{min}^*$ for PDS sampling pattern.

\begin{lemma}
\label{rmin_r}
As dimension $d$ approaches infinity, the maximum achievable relative ${r}_{min}^*$ converges to a constant, i.e.,
$$\lim_{d\rightarrow \infty} \hat{r}_{min}^* = \frac{1}{\sqrt{2\pi e}} $$
and, the minimum number of PDS samples needed to achieve $\hat{r}_{min}^*$ goes to infinity.
\end{lemma}

The results in Lemmas~\ref{rho_r} and \ref{rmin_r} show interesting limiting behaviors of both blue noise and PDS sampling distributions. 

\end{document}